\pdfoutput=1

\documentclass[11pt]{article}

\usepackage{ACL2023}
\usepackage{times}
\usepackage{ragged2e}
\usepackage{latexsym}
\usepackage{graphicx}
\usepackage{array}
\usepackage{hyperref}
\usepackage[normalem]{ulem}
\usepackage[T1]{fontenc}
\usepackage{mathtools}
\usepackage{adjustbox}
\usepackage{xcolor}
\usepackage{soul}
\usepackage{amsmath}
\definecolor{MyTurquoise}{RGB}{175,238,238} 
\definecolor{MyRed}{RGB}{255,204,203}
\definecolor{MyYellow}{RGB}{255,255,102}

\usepackage[utf8]{inputenc}

\usepackage{microtype}

\usepackage{inconsolata}
\usepackage[shortlabels]{enumitem}
\usepackage{tabularx, booktabs, subcaption, multirow}
\usepackage{longtable, multicol}
\usepackage{footmisc}

\title{ LLM for Comparative Narrative Analysis
 }

\author{Leo Kampen\textsuperscript{1}, Carlos Rabat Villarreal\textsuperscript{2}, Louis Yu\textsuperscript{1}, Santu Karmaker\textsuperscript{3},   Dongji Feng\textsuperscript{1}\\
\textsuperscript{1}  MCS department, Gustavus Adolphus College
\\
\textsuperscript{2} Department of CSSE, Auburn University\\
\textsuperscript{3} Bridge-AI Lab, Department of CS, University of Central Florida
\\
\{lkampen,djfeng\}@gustavus.edu}

\begin{document}
\maketitle

\begin{abstract}



In this paper, we conducted a Multi-Perspective Comparative Narrative Analysis (CNA) on three prominent LLMs: GPT-3.5, PaLM2, and Llama2. We applied identical prompts and evaluated their outputs on specific tasks, ensuring an equitable and unbiased comparison between various LLMs.
Our study revealed that the three LLMs generated divergent responses to the same prompt, indicating notable discrepancies in their ability to comprehend and analyze the given task. Human evaluation
was used as the gold standard, evaluating four perspectives to analyze differences in LLM performance.\footnote{https://github.com/Anonymous-685/Narrative-Sum-Project/tree/main/\label{fn:Github}}
\newcounter{myfoot}\setcounter{myfoot}{\value{footnote}}




\end{abstract}

\section{Introduction}\label{sec:intro}

In this paper, we formalized a Comparative Narrative Analysis (CNA) and provided an exploratory case study to better understand political narratives. CNA is a significant, yet relatively underexplored, task in the field of NLP, especially in the era of LLMs. It involves the challenging endeavor of identifying the summarization of 1) conflicts, 2) overlaps, 3) unique, or 4) holistic aspects across pairs of narratives.
Although prior research has explored various prompt designs to steer LLMs toward specific tasks~\cite{yao2022react,wei2022chain,kojima2022large,madaan2022text,press2022measuring}, benchmarking and comparing performance across multiple LLMs remains a significant challenge due to the variability introduced by differing prompt structures. 
To address this research gap, we utilized the recently proposed prompt taxonomy~\citet{santu2023teler}, TELeR, which organizes prompts along four dimensions: Turn, Expression, Role, and Level of Detail. The objective of our study is to apply identical prompts to different LLMs to evaluate their performance, ensuring consistency, reliability, and a fair comparison across a range of tasks.


However, evaluating such complex tasks is a challenge because applying traditional metrics like BERTScore~\cite{zhang2019bertscore} or BARTScore~\cite{yuan2021bartscore} may obscure the nuanced impact of specific perspectives, such as conflict or unique. Furthermore, the absence of a gold standard for evaluation exacerbates these challenges. 
Thus,  20 human evaluators were used to determine the effectiveness of LLM outputs across four widely established criteria.
To conclude our work,  this paper presents the experimental results of three LLMs evaluated using four levels of prompts based on the TELeR taxonomy across the summarization subtasks (subtask details in Section \ref{sec:ProblemDesign}).

\section{Background and Related Work}\label{sec:related}

\noindent\textbf{Prompt Design:} 
A prompt for an LLM is a collection of guidelines that can direct the Language Model toward a particular task ~\cite{liu2023pre}.
Currently, different prompt design strategies are proposed to improve the performance of LLMs in terms of such reasoning and decision-making ability ~\cite{zhou2022least}. The response of LLMs can vary significantly based on the prompt's quality. This variation can be attributed to different LLMs' diverse training datasets and annotations. Therefore, utilizing the same prompt to evaluate or explain multiple LLMs is advisable. To facilitate a meaningful comparison of the performances of various LLMs, Santu and Feng~(\citeyear{santu2023teler}) proposed a general taxonomy of LLM prompts for benchmarking complex tasks. 


\textbf{Multi-Perspective Narrative Analysis:} 
NLP techniques and multiple-document summarization ~\cite{CST} have been widely used in different tasks in the political domain ~\cite{political1, political2, political3} such as classification (e.g., fake news detection and author identification). Multiple perspective narrative analysis is similar to text summarization but further extends to analytical and interpretive tasks that can be injected into many applications such as a peer-reviewing generation where the meta-review can be quickly generated from reviews. 

\section{Problem Formulation}\label{sec:ProblemDesign}

\subsection{Comparative Narrative Analysis Task}

Although prior research~\cite{CST} has introduced multi-document summarization, a generalizable framework for formalizing this task remains lacking. Here, we introduce Comparative Narrative Analysis (CNA), generally broader as it involves not only summarizing content but also analyzing, contrasting, and interpreting narratives across multiple sources. 
To simplify notation, let us stick to having only two narratives $N^{}_{1}$ and $N^{}_{2}$ as our input since it can be easily generalized in case of more narratives involved. The desired output, denoted as $N^{}_{o}$, is a natural language summary where the format can be extractive summaries, abstractive summaries, or a combination of both.~\cite{karmaker2018sofsat,salvador2024benchmarking}
Mathematically, we can represent the CNA task as follows: 
\begin{equation}
N_o = \mathcal{F}(O, C, H, U)
\end{equation}

\noindent
where:
\begin{itemize}
    \item $N_o$ is the overall summary of the two narratives $N_1$ and $N_2$.
    \vspace{-2mm}
    \item $\mathcal{F}$ is a function that integrates the four perspectives.
    \vspace{-2mm}
    \item $O$, $C$, $H$, $U$ represent the \textbf{O}verlap, \textbf{C}onflict, \textbf{H}olistic, and \textbf{U}nique aspects of narratives respectively.


\end{itemize}

We can define $\mathcal{F}$ as a weighted sum shown below:  

\begin{equation}
N_o = \alpha O + \beta C + \gamma H + \delta U
\end{equation}

\noindent
where:
\begin{itemize}
    \item $\alpha, \beta, \gamma, \delta$ are weights that determine the contribution of each perspective to the overall summary.
\end{itemize}

In our study, each perspective is analyzed independently and exclusively, without considering the influence of other perspectives. However, the CNA framework has the potential for future expansion into a hybrid task.  As we can see,  this task can be framed as a constrained multi-sequence-to-sequence (text generation) problem, with an emphasis on minimizing verbatim repetition from the original narratives while incorporating analytical and interpretive elements.
While existing datasets such as Allsides~\cite{bansal2022semantic} facilitate research in overlapping subtasks, there is a lack of ground truth data for the other three perspectives (Holistic, Unique, Conflict). This necessitates the use of human evaluators to assess the quality of outputs generated by LLMs. Here are the detailed definitions of each CNA subtask: 1) \textbf{Holistic}: The generation fully highlights the information from two narratives without constraints, 2) \textbf{Overlapping}: The generation highlights the information that occurs in both narratives, 3)\textbf{Unique}: The generation highlights the distinctive information that only occurs in each narrative,
and 4) \textbf{Conflict}:  The generation highlights the opposing viewpoints or contradictory elements between two narratives.
Due to the page limit, we present the Venn diagram of the above four tasks in Appendix~\ref{appen:Venn}.

\subsection{TELeR Taxonomy}

TELeR taxonomy~\cite{santu2023teler} offers guidance for designing prompts at various levels.
 We designed our prompts using levels 1, 2, 3, and 4 (Find details in Appendix~\ref{appen:teller}). 
We did not use level 0 due to its lack of directives as we require concrete direction to specify which subset of CNA is being evaluated. We also left out levels 5 and 6 because they use retrieval techniques that are not relevant to this experiment. Prompts for all levels and subtasks can be found in Appendix~\ref{appen:prompt_table}




\section{Experimental Design}\label{sec:Experiment}

\subsection{Dataset}Bansal et al.~(\citeyear{bansal2022semantic}) created the SOS dataset by crawling articles from AllSides.com\footnote{https://www.allsides.com/}, which consists of news articles that cover a total of 2,925 events, each having a minimum 'theme description' length of 15 words. We randomly selected 5 pairs of narratives in this work and reconstructed a dataset totaling 240 narrative summarizations (3 models × 5 narrative pairs × 4 prompt levels × 4 subtasks). Our contributed dataset can be found here: 
GitHub\footnotemark[\value{myfoot}]


    

\subsection{Large Language Models}

For this experiment, we selected three of the most popular commercial LLMs: OpenAI GPT-3.5~\cite{achiam2023gpt}, Google PaLM2~\cite{anil2023palm2technicalreport},~\cite{chowdhery2022palm}, and Meta Llama2~\cite{touvron2023llama}. We utilized their respective APIs. Table~\ref{tab:LLMs} summarizes the details of these three LLMs.

\begin{table}[!thb]\footnotesize
\centering
\begin{tabular}{|lllll|}
\hline
\multicolumn{2}{|l|}{LLM Family}  & \multicolumn{3}{l|}{ Model} \\ \hline
\multicolumn{2}{|l|}{OpenAI GPT-3.5}  & \multicolumn{3}{l|}{openai/gpt-3.5-turbo-chat}      \\ \hline
\multicolumn{2}{|l|}{Google PaLM2} & \multicolumn{3}{l|}{google/palm2-bard-chat}      \\ \hline
\multicolumn{2}{|l|}{MetaAI Llama2} & \multicolumn{3}{l|}{meta-llama/Llama-2-7b-chat-hf (7B)}      \\ \hline

\end{tabular}%

\vspace{-3mm}

\caption{Large language models studied in this paper.}
\label{tab:LLMs}
\end{table}

\subsection{Surveys, Criteria and Human Evaluator}


We create surveys that include four key dimensions for assessment~\cite{fabbri2021summeval}: 1) \textbf{Coherence}, 2) \textbf{Consistency}, 3) \textbf{Relevance}, and 4) \textbf{Fluency}. In addition to the above four dimensions, we also customized one additional criterion for each specific subtask: Conflict is assessed for \textbf{bias}, Unique is evaluated for \textbf{distinctiveness}, Holistic is judged for \textbf{comprehensiveness}, and Overlapping is rated for \textbf{convergence}. Details of each customized criteria can be found in Appendix \ref{appen:criteria}.

Twenty human evaluators were recruited from Prolific\footnote{https://app.prolific.com/}, compensated with a minimum hourly rate of \$8/hour. 
Prolific provides filtering criteria to select human evaluators. We chose evaluators who are native English speakers, hold a bachelor’s degree, and have an understanding of political narratives.
We presented them with pairs of political news narratives and their corresponding summaries generated by each LLM. They then assessed the summaries across the above four plus one key dimensions. 
To streamline evaluation, we used a multiple-choice format where evaluators rated their level of satisfaction for each dimension. These satisfaction levels were then mapped to corresponding numerical scores.
For each annotator, there were 12 surveys (We provide a sample survey\footnote{https://forms.gle/RaRuYEm24SbdoGyw6}) split up by subtask and LLM used. 

The total amount of data points we have is 24,000 coming from 5 pairs of narratives each with 4 prompt levels. For each prompt level, we have 5 multiple choice criteria questions across 4 subtasks, tested on 3 LLMs with each getting evaluated by 20 annotators. Thus, we have 5 (pairs of narratives) × 4 (prompt level) × 5 (criteria) × 4 (subtasks) × 3 (LLM models) × 20 (annotators) = 24,000.
Each annotator assessed 1200 (5 × 4 × 5 × 4 × 3) data points.
Criteria details can be found in~\ref{appen:criteria}.

\section{Results}\label{sec:results}

\subsection{Experimental Results}

\begin{figure}[!htb]
    \centering
    \includegraphics[width=1\linewidth]{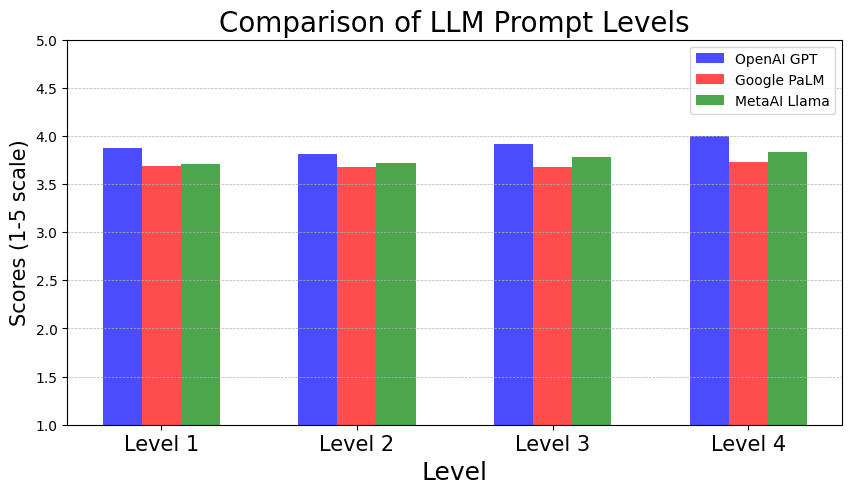}
    \caption{Comparison of different prompt levels} 
    \label{fig:combo_levels}
\end{figure}

\textbf{Prompt level 4 performs best on all three LLMs on average: }Figure~\ref{fig:combo_levels} illustrates the average results for the three LLMs across various subtasks and evaluation criteria at different prompt levels. As observed in the figure, prompt level 4 consistently yielded the highest average scores based on human evaluation, with an average rating of 4.0 on GPT-3.5. Similar results were observed on the other two LLMs.  In Figure~\ref{fig:combo_levels}, it shows that prompt level 4 can help LLMs generate results that are preferred by humans, with a score of 3.73 for PaLM2 and 3.83 for Llama2. 

While increasing prompt complexity often improves LLM performance, our findings reveal some unexpected results. For GPT-3.5, prompt level 3 (score: 3.92) achieved the second-highest performance, but prompt level 1 (score: 3.88) outperformed prompt level 2 (score: 3.81). Similarly, PaLM2 exhibited its second-best performance with prompt level 1 (score: 3.69). Meanwhile, GPT-3.5 also achieved the best performance at prompt level 4 (score: 4.0) among the three LLMs.


\vspace{-1mm}

\begin{figure}[!htb]
    \centering
     \includegraphics[width=1\linewidth]{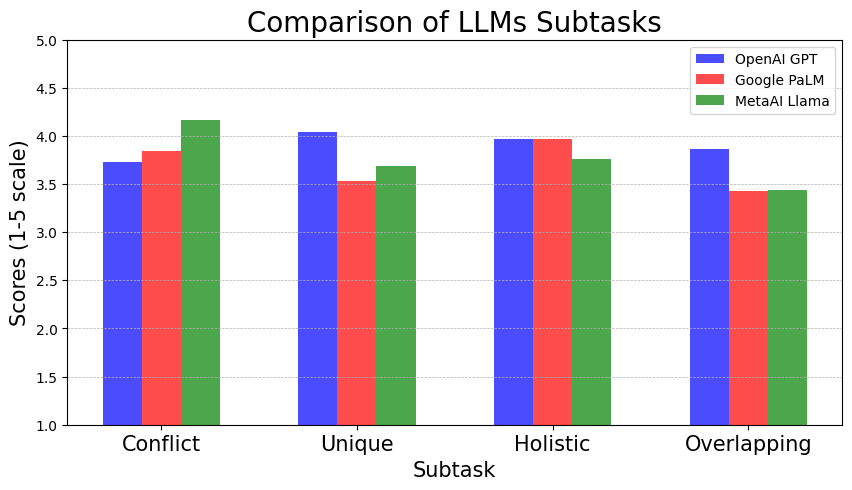}
    \caption{Comparison of subtasks across all levels and criteria}
    \label{fig:combo_sub}
    \vspace{-2mm}
\end{figure}

\textbf{LLMs exhibit varying performance across different subtasks:} Figure~\ref{fig:combo_sub} indicates a disparity in GPT-3.5's capabilities, excelling at identifying unique elements between narratives (score: 4.05) but struggling with conflict detection (score: 3.73). Llama2 demonstrates the strongest capability in identifying conflicts between narratives (score: 4.16), while struggling with the overlapping information subtask according to Figure~\ref{fig:combo_sub} (score: 3.44). This difficulty with overlap also appears to be the most challenging aspect for PaLM2 (score: 3.44) based on Figure~\ref{fig:combo_sub}. Notably, Llama2 outperforms both GPT-3.5 and PaLM2 on the conflict identification task.


\vspace{-1mm}

\begin{figure}[!htb]
    \centering
    \includegraphics[width=1\linewidth]{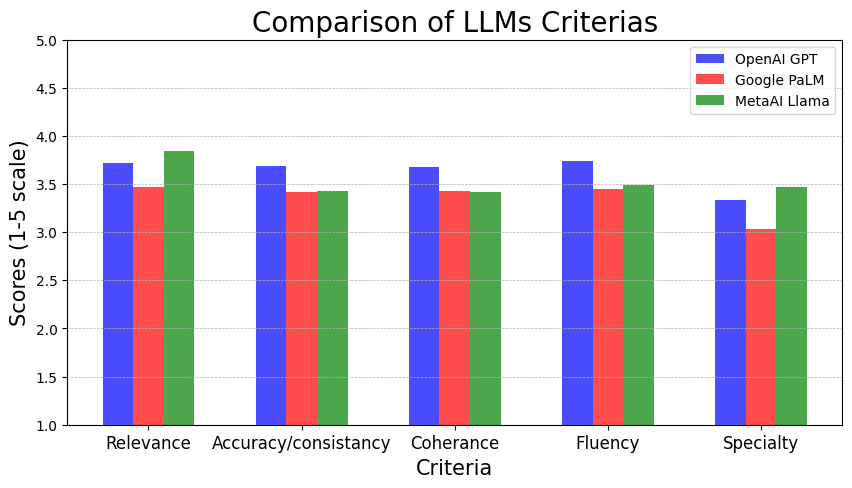}
    \caption{Comparison of criteria across all levels and subtasks}
    \label{fig:combo_crit}
\end{figure}

\textbf{Human expectations for LLMs vary based on specific criteria:} Figure~\ref{fig:combo_crit}
illustrates the average performance of GPT-3.5 on each evaluation criterion, including one subtask-specific "special dimension." While GPT-3.5 achieves a respectable average score of 3.7 across the four core dimensions, its performance on the special dimension falls short, with an average score of only 3.34.  Figure~\ref{fig:combo_crit} reveals that PaLM2's performance while achieving an average score within the expected range based on human evaluation criteria, falls short in the area of the specialty.  Meanwhile, Llama2 exhibits an interesting performance pattern on its evaluation criteria. While its score on the "relevance" criterion is significantly higher (3.84), the average score across the other three core dimensions was lower (3.45), according to Figure~\ref{fig:combo_crit}.

\vspace{-3mm}

\textbf{Statistical Significance Test:}
Here we examine whether there is a statistically significant difference in results from the following perspectives: 1)  models, 2)  prompt levels, 3)  evaluation criteria, and 4) subtasks. 
We first report ANOVA~\cite{st1989analysis, tavabi2024systematic} to give a comparing result among multiple models, then we provide specific pair-wise comparison from above perspectives using Tukey's HSD test~\cite{zhang2023exploring,abdi2010tukey}. Use the most widely used confidence value of 0.05 as the threshold, a p-value larger than 0.05 means the compared distributions are statistically the same, otherwise are statistically different~\cite{feng2023joint}. Table~\ref{tab:Statistical_table} demonstrates a statistically significant difference identified by the ANOVA analysis. It is evident that LLM generation varies across the subsets, such as different levels, leading to distinct outputs (p-value < 0.05), indicating a meaningful statistical difference.  More detailed comparisons are provided in Appendix~\ref{sec:appendix}.

\begin{table}
\centering
\resizebox{\columnwidth}{!}{%
\begin{tabular}{|l|l|l|l|l|}
\hline
        & Criteria & Level    & Subtasks & Model   \\ \hline
p-value & 4.20E-14 & 3.70E-05 & 4.5E-36  & 1.4E-14 \\ \hline
\end{tabular}%
}
\caption{ANOVA induced statistically significant differences from 1) criteria, 2) prompt level, 3) subtasks, and 4) each Model Perspective }
\label{tab:Statistical_table}
\end{table}

\begin{figure}[!htb]
    \centering
    \includegraphics[width=1\linewidth]{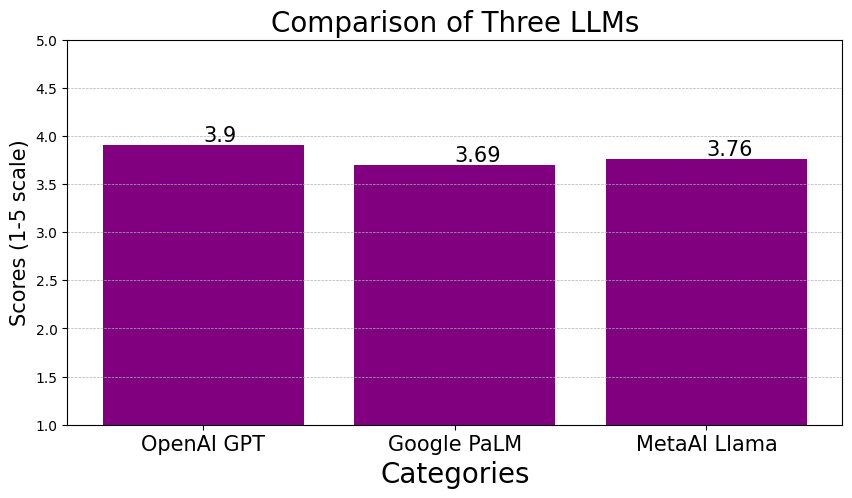}
    \caption{Comprehensive comparison of LLMs}
    \label{fig:Overall}
    \vspace{-1mm}
\end{figure}

\textbf{GPT-3.5 outperformed the other two LLMs in terms of average performance across the tasks:} Figure~\ref{fig:Overall} summarizes these results. As observed in the figure, GPT-3.5 achieved the highest average score (3.9), followed by Llama2 and PaLM2, which both received scores of 3.76 and 3.69 respectively.

\section{Discussion and Conclusion}\label{sec:conclusion}

In this paper, we explore the use of LLMs to address CNA tasks and present a case study on political narrative analysis.
Our findings reveal that prompt level 4 consistently led to the best performance across three LLMs, contrasting with prior studies relying on automated metrics~\cite{salvador2024benchmarking}. Task-specific analysis showed GPT-3.5 excelled in identifying unique elements, PaLM2 performed best on holistic tasks, and Llama2 was strongest in conflict identification. However, PaLM2 and Llama2 struggled with overlapping information tasks. While all models achieved average scores on general evaluation criteria, human evaluators expressed dissatisfaction with task-specific performance, with GPT-3.5 generally leading, followed by Llama2 and PaLM2.
In summary, this work paves the way for further research exploring LLMs' capabilities in CNA.





\section{Limitations}\label{sec:limitation}

Our study acknowledges three key limitations:

\textbf{Dataset Size:} The current dataset comprises only five narrative pairs limiting the generalizability of our findings.
However, the total number of data points gathered was calculated as follows: 4 subtasks × 3 LLM's × 5 narratives × 4 prompt levels × 5 criteria × 20 annotators = 24,000 annotations.
Thus, utilizing a larger and more diverse dataset in future work would allow for a more robust evaluation.

\textbf{Limited LLM Testing}: The evaluation was conducted on three LLMs. While this initial exploration demonstrates the efficacy of the TELeR taxonomy, including a broader range of LLMs in future studies would provide a more comprehensive understanding of its generalizability.


\textbf{Evaluation Limitations}: We created surveys because it is the best option behind a "Gold Standard" to compare the generated summaries against. "Gold Standards" are the ideal output for a given summary which takes resources we don't have available to us. Because of this, our research is relatively limited but future works will look to create a "Gold Standard" for each subtask to create automatic evaluation metrics.

Despite these limitations, our work offers valuable contributions:

\textbf{TELeR Taxonomy Validation:} This study successfully demonstrates the TELeR taxonomy's utility in establishing a fair and consistent evaluation framework for comparing LLM performance in complex NLP tasks like CNA.

\textbf{Human Evaluation Emphasis:} Our research underscores the importance of human evaluation, particularly for tasks requiring nuanced understanding like CNA. The extensive human evaluation (24,000 annotations) using diverse prompts and criteria strengthens the foundation for future research.

\bibliography{custom}
\bibliographystyle{acl_natbib}
\section{Ethics Statement}

In this paper, we have discussed behavior of LLMs while prompting with the TELeR taxonomy. Through this, we hope to assist new research directions. To the fulfillment of this goal, we have worked with real-world datasets. We did not obtain any explicit approval as our intended contents were already published for educational/research purposes. We have not tried to identify any private information from the data in any way which can result in a privacy violation. Additionally, the data we used (publicly released) does not contain personal information (e.g., usernames of users). In the whole experiment, we only used open source packages and libraries, along with proper citations as required also in accordance with its acceptable use policy, and no additional permission was required.

\newpage
\appendix

\section{Appendix}
\label{sec:appendix}

\subsection{Venn Diagram of subtasks}\label{appen:Venn}

Below is the Venn Diagram of four subtasks:

\begin{figure}[!htb]
    \centering
    \includegraphics[width=0.5\linewidth]{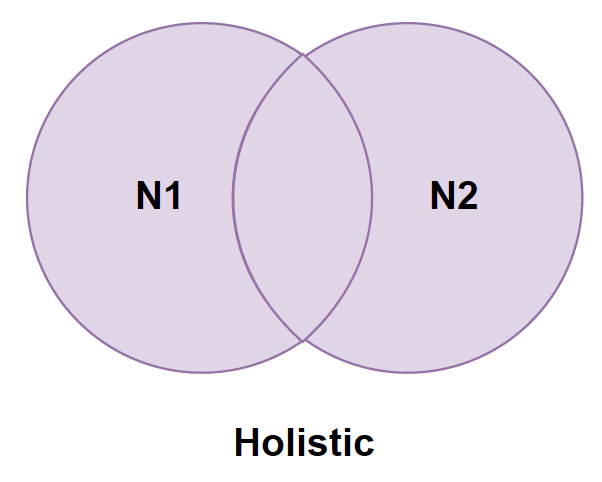}
    \caption{Holistic aspect of two narratives}
    \label{fig:Venn_Holistic}
\end{figure}

\begin{figure}[!htb]
    \centering
    \includegraphics[width=0.5\linewidth]{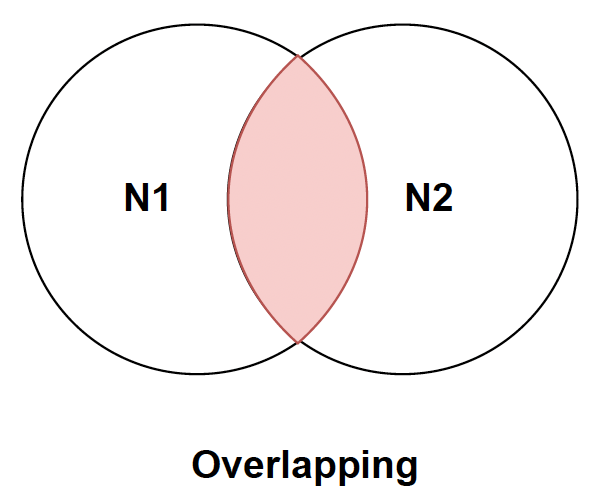}
    \caption{Overlapping aspect of two narratives}
    \label{fig:Venn_overlap}
\end{figure}

\begin{figure}[!htb]
    \centering
    \includegraphics[width=0.5\linewidth]{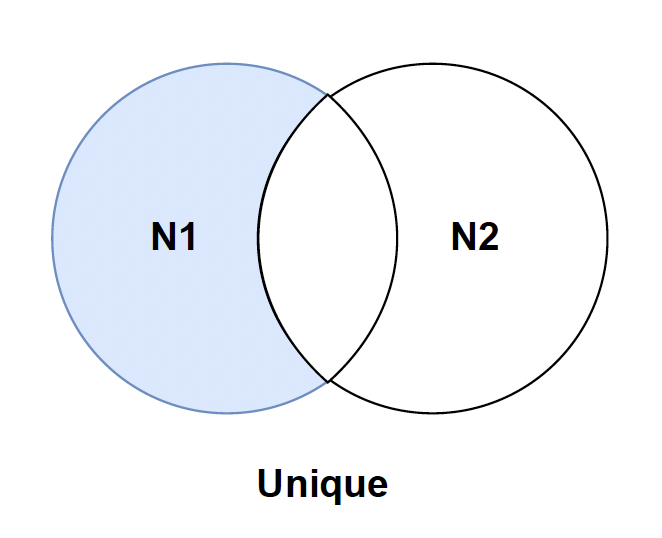}
    \caption{One of unique aspect of two narratives}
    \label{fig:Venn_unique1}
\end{figure}

\begin{figure}[!htb]
    \centering
    \includegraphics[width=0.5\linewidth]{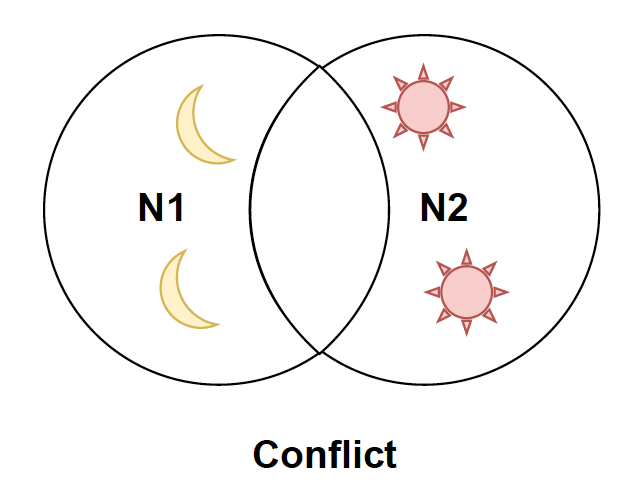}
    \caption{Conflict aspect of two narratives}
    \label{fig:Venn_conflict}
\end{figure}

\newpage








    

\subsection{General Criteria and Customized Criteria Definition}\label{appen:criteria}

\subsubsection{Human Evaluation Perspectives}

\textbf{Coherence} - the collective quality of all sentences. We align this dimension with the DUC quality question \cite{dang2005overview} of structure and coherence whereby "the summary should be well-structured and well-organized. The summary should not just be a heap of related information, but should build from sentence to sentence to a coherent body of information about a topic.

\textbf{Consistency} - the factual alignment between the summary and the summarized source. A factually consistent summary contains only statements that are entailed by the source document. Annotators were also asked to penalize summaries that contained hallucinated facts.

\textbf{Fluency }- the quality of individual sentences. Drawing again from the DUC quality guidelines, sentences in the summary "should have no formatting problems, capitalization errors or obviously ungrammatical sentences (e.g., fragments, missing components) that make the text difficult to read."

\textbf{Relevance} - selection of important content from the source. The summary should include only important information from the source 
\vspace{2mm}

In addition to the above four dimensions, we also customized one additional criterion for each specific subtask: Conflict is assessed for bias, Unique is evaluated for distinctiveness, Holistic is judged for comprehensiveness, and Overlapping is rated for convergence. Details of each customized criteria can be found below:

\textbf{Bias:}
The summary should not have a predisposition of a particular perspective. The summary should not include selected facts or language choices that favor one side or diminish the other.

\textbf{Distinctiveness:}
Does the summary highlight the distinct arguments in each narrative that are not found in the opposing one.

\textbf{Comprehensiveness:}
Does the summary cover all the essential points and not leave out crucial information. Does the summary include both the main points of both as well as individual points.

\textbf{Convergence:}
Does the summary highlight the parts of both narratives that agree.

\subsubsection{Human Evaluation Satisfaction Mapping}\label{sec:evaluatorMapping}

Table~\ref{tab:satisfying_level} shows our mapping function from human satisfaction levels to numerical scores. 

\begin{table}[!thb]\footnotesize
\resizebox{\columnwidth}{!}{%
\begin{tabular}{|l|c|}
\hline
Satisfaction Level  & Assigned Score \\ \hline
SA (Strongly Agree)    & 5              \\ \hline
A (Agree)           & 4              \\ \hline
N (Neutral)         & 3              \\ \hline
D (Disagree)        & 2              \\ \hline
SD (Strongly Disagree) & 1              \\ \hline
\end{tabular}%
}
\caption{Mapping Satisfaction Levels to Scores for Human Evaluation}
\label{tab:satisfying_level}
\end{table}





\subsection{Pairwise Statistical Significant Difference} \label{appen:pairwise_sta}

We present the results of statistical significance testing from four perspectives: 1) Models, 2) Subtasks, 3) Prompt levels, and 4) Criteria. Table~\ref{tab:Model_comparision} shows the comparisons between LLM models. All comparisons yield p-values less than 0.05 (rounded to 0 for very small values), indicating statistically significant differences.

\begin{table}[!thb]\footnotesize
\centering
\resizebox{0.8\columnwidth}{!}{%
\begin{tabular}{|l|l|l|}
\hline
Group 1 & Group 2 & P-Value \\ \hline
Palm2   & LLama   & 0.0281  \\ \hline
Palm2   & gpt3.5  & 0       \\ \hline
LLama   & gpt3.5  & 0       \\ \hline
\end{tabular}%
}
\caption{Model level Comparison }
\label{tab:Model_comparision}
\end{table}

Table~\ref{tab:prompt_level_comparision} presents the comparison across different prompt levels. An interesting observation from this table is that levels 1, 2, and 3 perform similarly, as indicated by p-values greater than 0.05, suggesting no statistically significant differences among them. In contrast, levels 3 and 4 exhibit significantly different performance.

\begin{table}[!thb]\footnotesize
\centering
\resizebox{0.8\columnwidth}{!}{%
\begin{tabular}{|l|l|l|}
\hline
Group 1 & Group 2 & P-Value \\ \hline
level 1 & level 2 & 0.86    \\ \hline
level 1 & level 3 & 0.84    \\ \hline
level 1 & level 4 & 0.0019  \\ \hline
level 2 & level 3 & 0.3792  \\ \hline
level 2 & level 4 & 0.001   \\ \hline
level 3 & level 4 & 0.0281  \\ \hline
\end{tabular}%
}
\caption{Prompt Level  Comparison }
\label{tab:prompt_level_comparision}
\end{table}

In table~\ref{tab:Subtasks_level_comparision},  we observe a significant difference between pairs of subtasks, except for the <conflict, holistic> pair. 

\begin{table}[!thb]\footnotesize
\centering
\resizebox{0.8\columnwidth}{!}{%
\begin{tabular}{|l|l|l|}
\hline
Group 1     & Group 2     & P-Value \\ \hline
conflict    & holistic    & 0.934   \\ \hline
conflict    & overlapping & 0       \\ \hline
conflict    & unique      & 0       \\ \hline
holistic    & overlapping & 0       \\ \hline
holistic    & unique      & 0       \\ \hline
overlapping & unique      & 0       \\ \hline
\end{tabular}%
}
\caption{Subtasks Level  Comparison }
\label{tab:Subtasks_level_comparision}
\end{table}

In the final pair-wise comparison, we demonstrate that different evaluation criteria yield varying results, except for the pairs <accuracy, relevance> and <coherence, relevance>, which show no significant difference, as shown in table~\ref{tab:Criteria__level_comparision}.

\begin{table}[!thb]\footnotesize
\centering
\resizebox{0.8\columnwidth}{!}{%
\begin{tabular}{|l|l|l|}
\hline
Group 1   & Group 2   & P-Value \\ \hline
accuracy  & coherence & 0.0058  \\ \hline
accuracy  & fluency   & 0.0047  \\ \hline
accuracy  & relevance & 0.051   \\ \hline
coherence & fluency   & 0       \\ \hline
coherence & relevance & 0.95    \\ \hline
fluency   & relevance & 0       \\ \hline
\end{tabular}%
}
\caption{Criteria Level  Comparison }
\label{tab:Criteria__level_comparision}
\end{table}

\newpage

\subsection{Survey Example}

This is a portion of one of our 12 surveys that we used for our human evaluation on our Comparative narrative analysis. The reason we did 12 surveys and split it up by subtasks and LLMs is because we have so much information to go through and a lot of reading for the evaluator. Because of this we split it up into smaller sections with each survey so we could then go in later and put all the data together which is where we got our results from.

\label{appen:Questionnair}

\begin{figure}[!htb]
    \centering
    \includegraphics[width=1\linewidth]{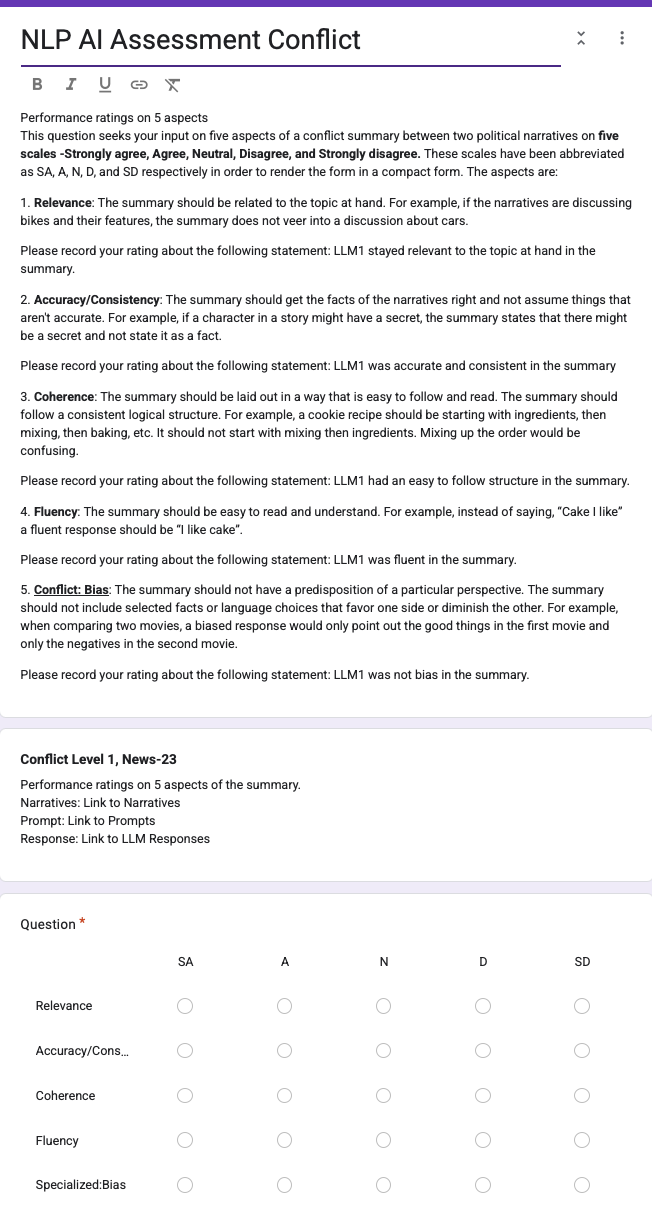}
    \caption{An example survey for a conflict summarization}
    
\end{figure}

\newpage
\subsection{TELeR Taxonomy}\label{appen:teller}

\begin{figure*}[!htb]
    \centering
\includegraphics[width=0.9\linewidth]{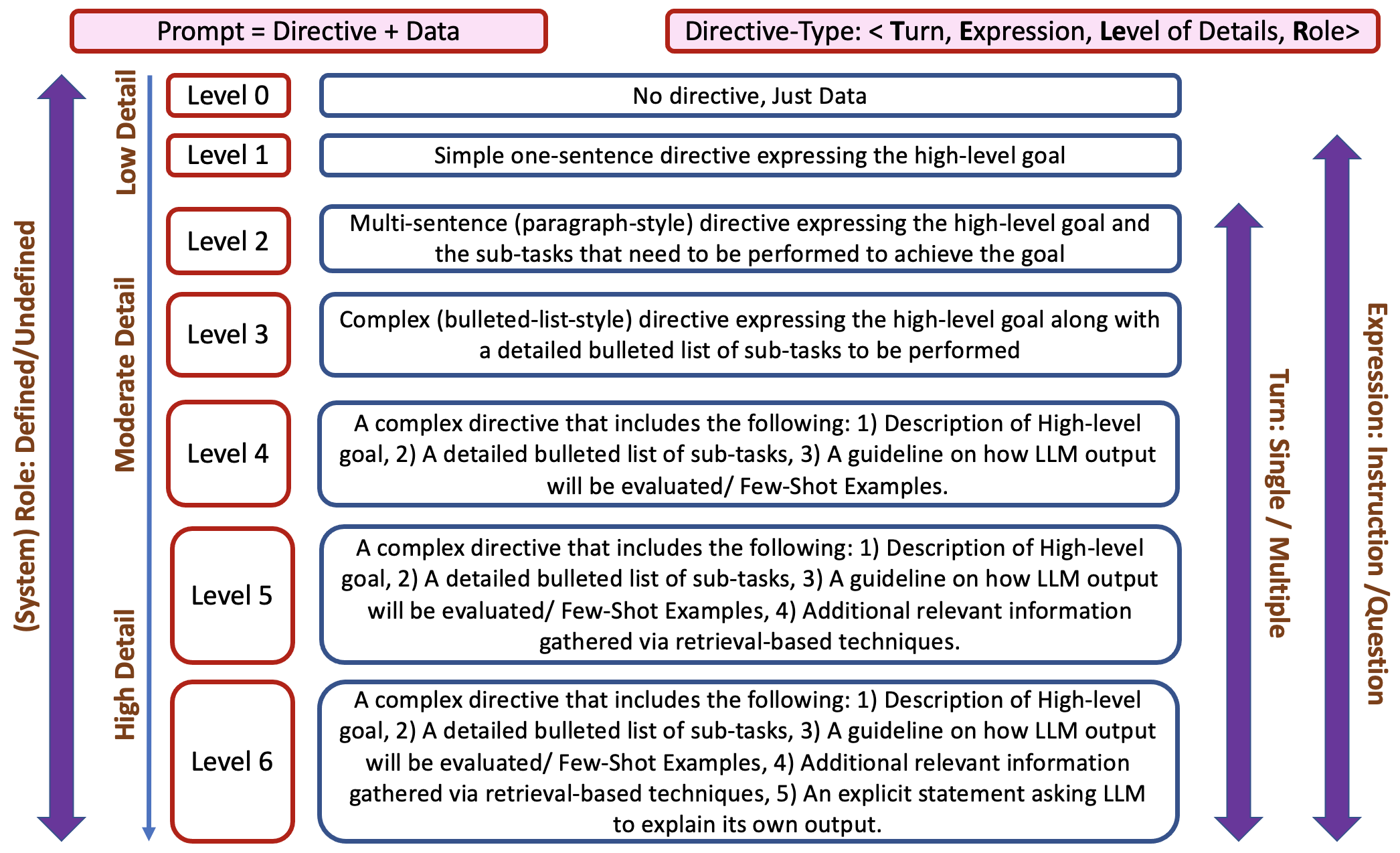}
    \vspace{-4mm}
    \caption{\textbf{TELeR} Taxonomy proposed by~\citet{santu2023teler}: (<\textbf{T}urn, \textbf{E}xpression, \textbf{Le}vel of Details, \textbf{R}ole>)}
    \label{fig:prompts_taxonomy}
    \vspace{-2mm}
\end{figure*}

Due to the significant variations in LLM performance when using different prompt types, styles, and levels of detail,~\citet{santu2023teler} 
proposes a general taxonomy for designing prompts with specific properties. This taxonomy aims to facilitate the effective execution of a wide range of complex tasks, see figure~\ref{fig:prompts_taxonomy}.

\newpage

\subsection{Prompt Example}\label{appen:prompt_table}

We provide detailed prompts designed according to the TELeR taxonomy for our subtasks Conflict, Holistic, Overlapping, and Unique across four levels in the table below.

\onecolumn

 \begin{longtable}{|p{0.1\textwidth}|p{0.8\textwidth}|}
 \hline
 \multicolumn{2}{|c|}{\textbf{Conflict}} \\ \hline
 Level 1 & Create a conflict summary between these two news stories where you highlight how the two news stories conflict with each other. \\ \hline
 Level 2 & Create a conflict summary between these two news stories where you highlight how the two news stories conflict with each other. The summary should include the opposing perspectives on the same issue and include the motivations for these different arguments. The summary should also stay concise and coherent throughout. \\ \hline
 Level 3 & Create a conflict summary between these two news stories where you highlight how the two news stories conflict with each other. The summary should include the following:
 \begin{itemize}
     \item An analysis of each side's perspectives and how they differ from each other
     \item An analysis of the motivations of the specific opinion of each side
     \item Specific differences of each side on the same issue
     \item Provide reasoning of where you got the information and why it is being included
 \end{itemize} \\ \hline
 Level 4 & Create a conflict summary between these two news stories where you highlight how the two news stories conflict with each other. The summary should include the following:
 \begin{itemize}
     \item An analysis of each side's perspectives and how they differ from each other
     \item An analysis of the motivations of the specific opinion of each side
     \item Specific differences of each side on the same issue
     \item Provide reasoning of where you got the information and why it is being included
 \end{itemize}
 In addition to the above, you will be evaluated on the following:
 \begin{itemize}
     \item Relevance: The summary should be related to the topic at hand.
     \item Accuracy/Consistency: The summary should get the facts of the narratives right and not assume things that aren't accurate.
     \item Coherence: The summary should be laid out in a way that is easy to follow and read. The summary should follow a consistent logical structure.
     \item Fluency: The summary should be easy to read and understand.
     \item Bias: The summary should not have a predisposition of a particular perspective. The summary should not include selected facts or language choices that favor one side or diminish the other.
 \end{itemize} \\ \hline
 \multicolumn{2}{|c|}{\textbf{Holistic}} \\ \hline
 Level 1 & Create a Holistic summary between these two news stories where you highlight the overall topic and message of the two news stories. \\ \hline
 Level 2 & Create a Holistic summary between these two news stories where you highlight the overall topic and message of the two news stories. The summary should include an overview of the topic that is being discussed in both narratives. There should be information pulled from both stories individually as well as parts where both stories overlap. \\ \hline
 Level 3 & Create a Holistic summary between these two news stories where you highlight the overall topic and message of the two news stories. The summary should include the following:
 \begin{itemize}
     \item An overview of the topic being discussed
     \item Information from both narratives individually
     \item Information that aligned with the other on the same issue
     \item Provide reasoning of where you got the information and why it is being included
 \end{itemize} \\ \hline
 Level 4 & Create a Holistic summary between these two news stories where you highlight the overall topic and message of the two news stories. The summary should include the following:
 \begin{itemize}
     \item An overview of the topic being discussed
     \item Information from both narratives individually
     \item Information that aligned with the other on the same issue
     \item Provide reasoning of where you got the information and why it is being included
 \end{itemize}
 In addition to the above, you will be evaluated on the following:
 \begin{itemize}
     \item Relevance: The summary should be related to the topic at hand.
     \item Accuracy/Consistency: The summary should get the facts of the narratives right and not assume things that aren't accurate.
     \item Coherence: The summary should be laid out in a way that is easy to follow and read. The summary should follow a consistent logical structure.
     \item Fluency: The summary should be easy to read and understand.
     \item Comprehensive: Does the summary cover all the essential points and not leave out crucial information. Does the summary include both the main points of both as well as individual points.
 \end{itemize} \\ \hline
 \multicolumn{2}{|c|}{\textbf{Overlapping}} \\ \hline
 Level 1 & Create an Overlapping summary between these two news stories where you highlight what the two news stories have in common. \\ \hline
 Level 2 & Create an Overlapping summary between these two news stories where you highlight what the two news stories have in common. The summary should include the aspects of each narrative that are the same as the other. It should only include shared information and opinions and none that are specific to an individual narrative. \\ \hline
 Level 3 & Create an Overlapping summary between these two news stories where you highlight what the two news stories have in common. Your summary should include the following:
 \begin{itemize}
     \item An analysis of shared information and opinions that are the same in both narratives
     \item Should not include information that is in one narrative but not the other
     \item Provide reasoning of where you got the information and why it is being included
 \end{itemize} \\ \hline
 Level 4 & Create an Overlapping summary between these two news stories where you highlight what the two news stories have in common. Your summary should include the following:
 \begin{itemize}
     \item An analysis of shared information and opinions that are the same in both narratives
     \item Should not include information that is in one narrative but not the other
     \item Provide reasoning of where you got the information and why it is being included
 \end{itemize}
 In addition to the above, you will be evaluated on the following:
 \begin{itemize}
     \item Relevance: The summary should be related to the topic at hand.
     \item Accuracy/Consistency: The summary should get the facts of the narratives right and not assume things that aren't accurate.
     \item Coherence: The summary should be laid out in a way that is easy to follow and read. The summary should follow a consistent logical structure.
     \item Fluency: The summary should be easy to read and understand.
     \item Convergent: Does the summary highlight the parts of both narratives that agree.
 \end{itemize} \\ \hline
 \multicolumn{2}{|c|}{\textbf{Unique}} \\ \hline
 Level 1 & Create a Unique summary between these two news stories where you highlight where the two news stories have information or opinions that are not in the other. \\ \hline
 Level 2 & Create a Unique summary between these two news stories where you highlight where the two news stories have information or opinions that are not in the other. The summary should include only the information that is unique to each narrative and should have no information or opinions that are in both. \\ \hline
 Level 3 & Create a Unique summary between these two news stories where you highlight where the two news stories have information or opinions that are not in the other. The summary should include the following:
 \begin{itemize}
     \item An analysis of the information and opinions that are in one narrative but not the other
     \item Should not include information that is in both narratives, only information that is unique
     \item Provide reasoning of where you got the information and why it is being included
 \end{itemize} \\ \hline
 Level 4 & Create a Unique summary between these two news stories where you highlight where the two news stories have information or opinions that are not in the other. The summary should include the following:
 \begin{itemize}
     \item An analysis of the information and opinions that are in one narrative but not the other
     \item Should not include information that is in both narratives, only information that is unique
     \item Provide reasoning of where you got the information and why it is being included
 \end{itemize}
 In addition to the above, you will be evaluated on the following:
 \begin{itemize}
     \item Relevance: The summary should be related to the topic at hand.
     \item Accuracy/Consistency: The summary should get the facts of the narratives right and not assume things that aren't accurate.
     \item Coherence: The summary should be laid out in a way that is easy to follow and read. The summary should follow a consistent logical structure.
     \item Fluency: The summary should be easy to read and understand.
     \item Distinct: Does the summary highlight the distinct arguments in each narrative that are not found in the opposing one.
 \end{itemize}  \\ \cline{1-2} 
\multicolumn{1}{p{0.1\textwidth}}{\colorbox{white}}
 \label{table:prompt_table}
 \end{longtable}

\twocolumn





\end{document}